\documentclass{article}
\usepackage{graphicx} 
\usepackage{booktabs}
\usepackage{amsmath}
\usepackage{url}
\usepackage{tablefootnote}
\usepackage{authblk}
\usepackage{siunitx}
\usepackage[table]{xcolor}
\usepackage{adjustbox}
\usepackage{caption}
\usepackage{hyperref}

\title{OpenJAI-v1.0: An Open Thai Large Language Model \\[1.0ex] 
    \normalsize 
    {%
        {\includegraphics[height=1.2em]{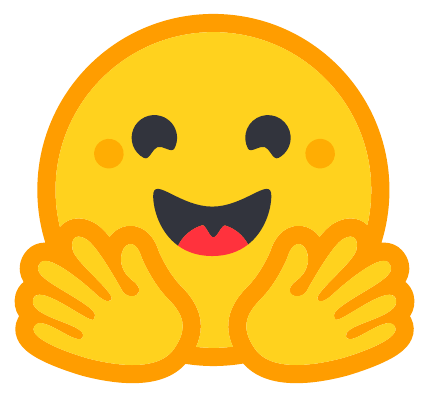}}%
        \href{https://huggingface.co/JTS-AI}{https://huggingface.co/JTS-AI} 
    }
}
\author[1]{Pontakorn~Trakuekul}
\author[1,2]{Attapol~T.~Rutherford}
\author[1]{Jullajak~Karnjanaekarin}
\author[1]{Narongkorn~Panitsrisit}
\author[1]{Sumana~Sumanakul}
\affil[1]{Jasmine Technology Solution} 
\affil[2]{Department of Linguistics, Chulalongkorn University} 
\affil[ ]{\texttt{jts.ai.team@gmail.com}}

\date{October 2025}

\begin{document}

\maketitle


\begin{abstract}

We introduce OpenJAI-v1.0, an open-source large language model for Thai and English, developed from the Qwen3-14B model. Our work focuses on boosting performance on practical tasks through carefully curated data across three key use cases: instruction following, long-context understanding, and tool use. Evaluation results show that OpenJAI-v1.0 improves on the capabilities of its base model and outperforms other leading open-source Thai models on a diverse suite of benchmarks, while avoiding catastrophic forgetting. OpenJAI-v1.0 is publicly released as another alternative NLP resource for the Thai AI community.

\end{abstract}

\section{Introduction}
The rapid advancement of Large Language Models (LLMs) has redefined the boundaries of artificial intelligence, yet this progress has been unevenly distributed. A notable performance disparity often exists between the English-language proficiency of state-of-the-art models and their capabilities in other languages, including Thai. This gap presents the necessity of developing high-quality, open-source foundation models to catalyze a robust AI ecosystem for the Thai language. To contribute to the growing ecosystem of Thai language models, which includes notable projects such as Typhoon \cite{typhoon2, typhoont1}, OpenThaiGPT \cite{openthaigpt1.5, openthaigpt1.6}, and WangchanLion \cite{mangosteen}, we introduce OpenJAI-v1.0, an open-source model developed to achieve competitive capabilities in Thai and English.

Our primary motivation is to build upon the strong linguistic foundation of a pretrained model to deliver deep practical utility. We selected the robust and efficient Qwen3-14B \cite{qwen3} as our base. Building on this foundation, our development methodology was deliberately oriented toward enhancing three core capabilities that define the usefulness of large language models in real-world applications: complex instruction following, which enables the model to reliably execute multi-step and nuanced user prompts; long-context understanding, which allows the model to maintain coherence and accuracy over extended inputs such as long documents, conversations, or legal texts; and reliable tool use, which equips the model to integrate seamlessly with external APIs, retrieval systems, and software environments. Together, these capabilities position the model as not only a linguistically competent system but also a practically deployable foundation for diverse Thai-language applications. We evaluated our model on benchmark datasets that reflect these capabilities in both Thai and English and demonstrated the improvement over some strong baselines.

This paper details the methodology behind OpenJAI-v1.0, from our data curation strategies to our comprehensive evaluation framework. We present a rigorous, multi-faceted evaluation that benchmarks OpenJAI-v1.0 against its base model and other leading open-source Thai models. To contribute to the open-source AI community, we are publicly releasing the OpenJAI-v1.0 model family to foster further research and application development.

\section{Data and Experimental Setup}
The strong performance of OpenJAI-v1.0 is primarily attributed to a meticulously curated, high-quality dataset spanning Thai and English. Notably, all samples are structured in a direct instruction-response format without explicit intermediate reasoning steps. The dataset is structured across three key domains:

\begin{itemize}
    \item \textbf{Instruction Following}: Our instruction-following dataset was constructed by curating high-quality public datasets and generating novel synthetic data. The synthesized samples were generated to strictly adhere to all predefined constraints and then rigorously filtered by an LLM-as-a-judge. The judge assessed each sample for constraint adherence, helpfulness, and alignment with user intent, rejecting any that failed to meet our quality standards. This crucial filtering step prevents the model from learning to satisfy constraints literally while diverging from the user's underlying objective. To ensure our evaluation truly reflects the model's effectiveness on real-world challenges, we excluded all constraints from the IFBench benchmark \cite{ifbench} during training, preserving a true zero-shot assessment of the model's generalization capabilities.

    \item \textbf{Long-Context Understanding}: Mastering long-context comprehension is essential for a wide range of applications. To enhance this capability, we constructed a robust dataset by combining curated, high-quality open-source data with our synthetic samples, specifically designed to improve the model's robustness and coherence when processing context lengths up to 120,000 tokens, making it highly effective for real-world Retrieval-Augmented Generation (RAG) tasks that involve extensive document sets.

    \item \textbf{Tool Calling}: Tool calling is a critical capability that enables an LLM to interpret natural language as executable actions by interacting with external tools and APIs. To develop this functionality, we curated a comprehensive dataset from established sources to encompass diverse scenarios, including both single- and multi-turn interactions. We also translated a selection of these datasets into Thai to ensure strong bilingual performance. Crucially, the dataset includes examples that require the model to discern whether to invoke a tool or respond conversationally. 
\end{itemize}

All model training and development were conducted on a dedicated 8xH100 GPU cluster in Jasmine Technology Solution's in-house GPU farm. The model was trained on approximately 462 million tokens. The global batch size was set to 256 instances. And the training process took less than one day. 

\section{Evaluation and Benchmarks}
We evaluated OpenJAI-v1.0 against a comprehensive suite of benchmarks to validate its performance across key capabilities. To ensure a standardized environment for all open-source models, we hosted them locally on our infrastructure, using the vLLM inference server \cite{vllm}. All evaluations were conducted in non-thinking mode, and all default system prompts were removed to ensure a fair comparison.

\subsection{Instruction Following}
For our evaluation, we report accuracy as the average of strict prompt-level and instruction-level accuracy. 

\begin{itemize}
    \item \textbf{IFBench-EN} \cite{ifbench}: This benchmark from the Allen Institute for AI was specifically designed to address a critical issue in model evaluation: benchmark overfitting. This problem is prevalent in older benchmarks, such as IFEval \cite{ifeval}, where a limited set of verifiable constraints allows models to be finetuned specifically for high performance without a genuine improvement in general instruction-following ability. IFBench mitigates this risk by providing a more diverse and challenging test suite. It comprises 300 English prompts with 58 novel, out-of-domain constraints, enabling a more accurate assessment of a model's true generalization capabilities.
 
    \item \textbf{IFBench-TH}: Given the lack of a comparable benchmark for Thai, we developed IFBench-TH by meticulously adapting its English counterpart. The process began with an in-house machine translation of the original IFBench prompts and arguments, which were then manually refined by our team to ensure accuracy and consistency. We adapted the benchmark to account for the unique characteristics of the Thai language, excluding constraints that were linguistically impractical. For example, tasks requiring precise sentence segmentation or the identification of sentence types based on punctuation (e.g., question or exclamation marks) were excluded, as these features are not standard in written Thai. This careful adaptation process resulted in a challenging yet fair benchmark containing a final set of 180 prompts.

\end{itemize}

\subsection{Multi-turn Capability}
\begin{itemize}
    \item \textbf{MT-Bench-EN} \cite{mtbench}: MT-Bench is a prominent benchmark for assessing the multi-turn conversational and instruction-following capabilities of LLMs. It comprises 80 challenging multi-turn questions across eight key domains: writing, roleplay, extraction, reasoning, math, coding, knowledge I (STEM), and knowledge II (humanities/social science). An LLM-as-a-judge evaluates model responses on a 10-point scale according to criteria such as helpfulness, relevance, and accuracy across both conversational turns.
    \item \textbf{MT-Bench-TH} \cite{mtbenchthai}: Evaluation of Thai capabilities was conducted using the culturally adapted version of MT-Bench from the ThaiLLM leaderboard. A key modification in this benchmark is the introduction of a novel category, Knowledge III, designed to evaluate comprehension of Thai cultural context. GPT-4o-2024-08-06 \cite{gpt4o} was employed as the LLM-as-a-judge for this evaluation. 
\end{itemize}

\subsection{Long-Context Understanding}
To ensure a fair long-context evaluation, we standardized the protocol by setting a uniform maximum input length of 120,000 tokens for all models. Consequently, models with native context windows shorter than this limit were adapted using Dynamic RoPE scaling for context extension.
\begin{itemize}
    \item \textbf{MRCR}\footnote{https://huggingface.co/datasets/openai/mrcr}: OpenAI's Multi-Round Coreference Resolution benchmark evaluates a model's ability to retrieve specific information from long, dense text. This needle-in-a-haystack task challenges the model to identify and differentiate multiple unique pieces of information ("needles") that are intentionally embedded within a lengthy document ("haystack") filled with similar distractor information. Our evaluation was conducted on the 8-needle variant of the benchmark.
    \item \textbf{LongBench-v2} \cite{longbenchv2}: A challenging multi-task benchmark designed to assess LLM performance on long-context problems requiring deep reasoning. It contains 503 multiple-choice questions with contexts ranging up to two million tokens, covering diverse and realistic tasks such as multi-document question answering, code repository comprehension, and analysis of long dialogues. The evaluation was conducted in non-CoT mode, and we report the overall score.

\end{itemize}

\subsection{Tool Calling}
For this evaluation, we utilized the native function-calling mode for each model and report the overall benchmark scores. To mitigate errors from context length limitations in multi-turn dialogues, Dynamic RoPE scaling was applied. This may cause minor discrepancies when comparing our results to those published by the original authors.

\begin{itemize}
    \item \textbf{BFCL-v3-EN} \cite{bfcl}: A comprehensive benchmark designed to evaluate an LLM's ability to invoke tools and functions in diverse real-world situations. It evaluates performance on tasks of varying complexity, including simple function calls, parallel execution of multiple functions, and intricate multi-turn dialogues that necessitate contextual memory and multi-step reasoning.

    \item \textbf{BFCL-v3-TH}: To enable tool-calling evaluation in a Thai context, our team developed BFCL-v3-TH. This specialized benchmark was created by applying in-house machine translation to the contents of the original English benchmark.

\end{itemize}

\subsection{General Knowledge}
We report the average score across all subjects for each model.
\begin{itemize}

    \item \textbf{MMLU-ProX-lite-EN} \cite{mmluprox}: We used this benchmark to verify the preservation of the base model's core knowledge post-finetuning. Its broad subject coverage serves as a crucial diagnostic to ensure our training methodology did not induce significant knowledge degradation.
    \item \textbf{MMLU-ProX-lite-TH}: The official Thai counterpart was used to assess the model's knowledge base and reasoning capabilities within the Thai language context. This benchmark is crucial for assessing whether the model's proficiency extends beyond translation to a genuine understanding of Thai-centric subjects.

\end{itemize}

\section{Results and Discussion}
The comprehensive evaluation results are presented in Table \ref{tab:model_benchmark}. We benchmarked the performance of OpenJAI-v1.0 against its base model (Qwen3-14B), other leading open-source Thai models of comparable size (Typhoon2.1-gemma3-12b and OpenThaiGPT1.5-14b), and a proprietary baseline (GPT-4.1-nano-2025-04-14). We note that the reported MT-Bench score for GPT-4.1-nano may exhibit a favorable bias, as the evaluation was conducted using GPT-4o as the judge—a model from the same family \cite{panickssery2024llm}.

\definecolor{lightpink}{HTML}{FCE5D7}
\definecolor{lightyellow}{HTML}{FEF2CB}
\definecolor{lightgreen}{HTML}{E2EFDA}
\definecolor{lightblue}{HTML}{DEEAF6}
\definecolor{headercolor}{HTML}{FFE699}
\definecolor{lightgray}{HTML}{E0E0E0}

\begin{table}[htbp]
\centering

\vspace{2pt} 

\resizebox{\textwidth}{!}{%
\begin{tabular}{
  l
  S[table-format=2.1, table-text-alignment=center]
  S[table-format=2.1, table-text-alignment=center]
  S[table-format=2.1, table-text-alignment=center]
  S[table-format=2.1, table-text-alignment=center]
  S[table-format=2.1, table-text-alignment=center]
}
\toprule
\rowcolor{white}
\textbf{Benchmark/Model} &
{\adjustbox{valign=c}{\textbf{\shortstack{OpenJAI-v1.0\\-14b}}}} &
{\adjustbox{valign=c}{\textbf{\shortstack{Qwen3\\-14b}}}} &
{\adjustbox{valign=c}{\textbf{\shortstack{Typhoon2.1-\\gemma3-12b}}}} &
{\adjustbox{valign=c}{\textbf{\shortstack{OpenThaiGPT1.5\\-14b}}}} &
{\adjustbox{valign=c}{\textbf{\shortstack{GPT-4.1-nano-\\2025-04-14}}}} \\
\midrule

\rowcolor{lightgray}
\multicolumn{6}{l}{\textbf{Instruction Following}} \\
IFBench-EN              & {\textbf{32.4}} & 29.7            & 27.4            & {\underline{30.6}}  & 28.3 \\
IFBench-TH              & {\textbf{39.4}} & {\underline{38.1}} & 36.5            & 35.4            & 34.9 \\
\addlinespace 

\rowcolor{lightgray}
\multicolumn{6}{l}{\textbf{Multi-turn Capability}} \\
MT-Bench-EN             & {\underline{8.4}} & {\underline{8.4}} & 8.3             & 7.8             & {\textbf{8.5}} \\
MT-Bench-TH             & {\textbf{8.1}} & {\underline{8.0}} & {\textbf{8.1}} & 6.9             & {\underline{8.0}} \\
\addlinespace

\rowcolor{lightgray}
\multicolumn{6}{l}{\textbf{Long-Context Understanding}} \\
MRCR           & {\textbf{18.9}} & {\underline{18.3}} & 16.9            & 16.9            & 16.2 \\
LongBench-v2            & {\textbf{33.6}} & 32.4            & 29.2            & {\textbf{33.6}} & 28.8 \\
\addlinespace

\rowcolor{lightgray}
\multicolumn{6}{l}{\textbf{Tool Calling}} \\
BFCL-v3-EN              & {\textbf{60.5}} & {\underline{59.2}} & 52.2            & 52.9            & 53.1 \\
BFCL-v3-TH              & {\textbf{47.0}} & {\underline{46.0}} & 45.0            & 44.9            & 41.1 \\
\addlinespace

\rowcolor{lightgray}
\multicolumn{6}{l}{\textbf{General Knowledge}} \\
MMLU-ProX-lite-EN       & {\underline{66.0}}            & {\textbf{66.6}} & 55.1            & 64.3            & 36.3 \\
MMLU-ProX-lite-TH       & {\underline{54.7}}            & {\textbf{57.5}} & 45.2            & 49.3            & 39.8 \\
\bottomrule
\end{tabular}
}

\caption{Benchmark results comparing OpenJAI-v1.0 (14B) with its base model (Qwen3-14B), other Thai-focused open-source models (Typhoon2.1-gemma3-12B, OpenThaiGPT1.5-14B), and a proprietary baseline (GPT-4.1-nano). Scores are reported across five capability categories: instruction following, multi-turn dialog, long-context understanding, tool calling, and general knowledge.}
\label{tab:model_benchmark}

\end{table}

The evaluation results presented in Table \ref{tab:model_benchmark} position OpenJAI-v1.0 as a leading open-source model for both Thai and English, consistently outperforming other prominent Thai models in key areas.


A primary area of advancement lies in instruction following. On IFBench, OpenJAI-v1.0 reaches 32.4 in English and 39.4 in Thai, improving over Qwen3-14B by 2.7 and 1.3 absolute points, respectively. These gains are particularly notable given that instruction following is a demanding capability where small absolute increases correspond to large differences in user-perceived quality. In both languages, OpenJAI-v1.0 outperforms other open-source Thai models and matches or surpasses the proprietary GPT-4.1-nano baseline. These outcomes underscore the effectiveness of our targeted post-training strategy.

Long-context reasoning further highlights the benefits of our finetuning approach. On MRCR, OpenJAI-v1.0 obtains 18.9, slightly higher than Qwen3-14B. On LongBench-v2, it reaches 33.6, outperforming all but one baseline, with which it ties. These findings suggest that the model has become more resilient when handling extended documents and retrieval-style inputs. Such resilience is crucial in production environments such as legal analysis, customer support, or educational applications.

The most substantial margins appear in tool calling. On BFCL-v3, OpenJAI-v1.0 records 60.5 in English and 47.0 in Thai. The model not only improves upon Qwen3-14B but also surpasses Typhoon2.1-gemma3-12b, OpenThaiGPT1.5-14b, and GPT-4.1-nano. Tool use represents a crucial capability for integrating LLMs into structured workflows. These results show that OpenJAI-v1.0 is particularly well suited for multi-step, tool-augmented applications.

Equally important is the preservation of broad factual knowledge. On MMLU-ProX-lite, OpenJAI-v1.0 maintains 66.0 in English and 54.7 in Thai. These scores closely track those of Qwen3-14B (66.6 and 57.5, respectively). Although there is a slight decline in Thai, the performance remains higher than that of competing Thai models. This outcome indicates that our specialization process did not induce catastrophic forgetting. General knowledge remains intact alongside the enhanced targeted capabilities.

Taken together, the results highlight a favorable balance between specialization and retention. OpenJAI-v1.0 leads or ties in the majority of evaluated dimensions, demonstrates bilingual improvements in core skills, and shows the strongest relative gains in areas of practical significance. Our experimental results confirm that the design of our training pipeline successfully yielded a model that is both broadly capable and optimized for the real-world use cases most relevant to Thai users. 

\section{Conclusion and Future Work}
OpenJAI-v1.0 marks a substantial contribution to the open-source AI ecosystem for the Thai language. By finetuning a powerful base model on high-quality, domain-specific datasets, we have produced a model that excels in practical, real-world capabilities while preserving its foundational knowledge in both Thai and English.

Our immediate roadmap includes expanding the OpenJAI family with additional model sizes to cater to different resource and performance requirements. Looking ahead, we plan to enhance our training data with more complex, culturally-specific Thai contexts and explore advanced techniques to broaden its range of practical, real-world applications.

\bibliographystyle{unsrt}
\bibliography{reference}

\end{document}